# From Likelihood to Plausibility

**Paul-André Monney**
University of Fribourg, Seminar of Statistics
Beauregard 11
CH-1700 Fribourg, Switzerland
paul-andre.monney@unifr.ch

## Abstract

Several authors have explained that the like-lihood ratio measures the strength of the evidence represented by observations in statistical problems. This idea works fine when the goal is to evaluate the strength of the available evidence for a simple hypothesis versus another simple hypothesis. However, the applicability of this idea is limited to simple hypotheses because the likelihood function is primarily defined on points — simple hypotheses — of the parameter space. In this paper we define a general weight of evidence that is applicable to both simple and composite hypotheses. It is based on the Dempster-Shafer concept of plausibility and is shown to be a generalization of the likelihood ratio. Functional models are of a fundamental importance for the general weight of evidence proposed in this paper. The relevant concepts and ideas are explained by means of a familiar urn problem and the general analysis of a real-world medical problem is presented.

## 1  The Problem

In a recent book, Richard Royall [11] explains that the likelihood ratio is the appropriate concept for measuring the strength of the evidence represented by observations in statistical problems. He gives several very convincing arguments in favor of this idea and shows its relevance in the field of testing statistical hypotheses. This interpretation of likelihood is based on Ian Hacking's *law of likelihood* [7] :

*If hypothesis A implies that the probability that a random variable X takes the value x is $P_A(x)$, while hypothesis B implies that the probability is $P_B(x)$, then the observation $X = x$ is evidence supporting A over B if and only if $P_A(x) > P_B(x)$, and the likelihood ratio,* $P_A(x)/P_B(x)$, *measures the strength of that evidence (I. Hacking, 1965).*

To elaborate on this idea we consider two spaces : the parameter space $\Theta$ and the observation space $X$. The set $\Theta$ is the set of possible values of the parameter variable $t$, whereas $X$ is the set of possible values of the observable variable $\xi$. There is exactly one correct but unknown value of the parameter variable $t$. Following Dawid & Stone [1], a distributional model is the specification of conditional distributions $P_\theta$ of the variable $\xi$ given $t = \theta$, i.e.

$$P(\xi = x | t = \theta) = P_\theta(x). \tag{1}$$

For each value $x$ in $X$, the function $l_x : \Theta \to [0,1]$ given by $l_x(\theta) = P_\theta(x)$ is the likelihood function of the observation $x$. According to the law of likelihood, the observation $X = x$ is evidence supporting the hypothesis $t = \theta$ over the hypothesis $t = \theta'$ if

$$l_x(\theta)/l_x(\theta') > 1 \tag{2}$$

and the weight of that evidence is precisely

$$w_x(\theta, \theta') := l_x(\theta)/l_x(\theta'). \tag{3}$$

This defines a function

$$w_x : \Theta \times \Theta \to [0, \infty] \tag{4}$$

which is called the weight of evidence function. Given a distributional model and an observation $x$, we can compute the weight of evidence of a simple hypothesis $H = \{\theta\}$ over a simple hypothesis $H' = \{\theta'\}$. But the weight of evidence is not defined for composite hypotheses (a hypothesis is composite if it contains more than one element). As Royall [11] says, in general the law of likelihood is silent for composite hypotheses. In the next section we are going to define a general notion of weight of evidence that is applicable to any kind of hypotheses. Of course, this general weight of evidence will be compatible with the likelihood ratio, i.e. when the general weight of evidence is applied to simple hypotheses the result is the likelihood ratio.

## 2    Likelihood and Plausibility

In Monney [9] the classical notion of a functional model introduced by Bunke [4] and studied by Dawid & Stone [1] has been generalized to the so-called generalized functional models (abbreviated by GFM in this paper). It turns out that these models can be used to give a sound and reasonable definition of a general weight of evidence. Distributional models don't contain enough information to be able to compute weights of evidence for composite hypotheses. We need to leave the realm of distributional models and enter the more informative class of generalized functional models to reach that goal.

A generalized functional model is a pair $(f, P)$ where $f$ is a function and $P$ is a probability distribution. If $\Omega$ denotes the set of possible values of a random variable $\epsilon$, then the function $f$ is a mapping $f : \Theta \times \Omega \to X$ with $f(\theta, \omega)$ representing the value of the variable $\xi$ that must necessarily be observed if $\theta$ is the correct value of the parameter variable $t$ and the realization of the random variable $\epsilon$ is $\omega$. The distribution of $\epsilon$ is the known probability measure $P$ on $\Omega$ and in a GFM the function $f$ is assumed to be known. By the way, note that a GFM can be seen as a particular causal theory in the sense of Pearl [10].

The observation of a value $x$ for the variable $\xi$ in a GFM generates some information about the unknown correct value of the parameter variable $t$. It turns out that this is a Dempster-Shafer belief function which will be denoted by $\mathcal{H}_x$ (Shafer [12], Kohlas & Monney [8]). More precisely, if

$$v_x(\Omega) = \{\omega \in \Omega : \exists \, \theta \in \Theta \text{ such that } f(\theta, \omega) = x\}, \tag{5}$$

then for every $\omega \in v_x(\Omega)$ there is a corresponding focal set of $\mathcal{H}_x$ given by

$$\Gamma_x(\omega) = \{\theta \in \Theta : f(\theta, \omega) = x\}. \tag{6}$$

The basic probability assignment (or $m$-function) $m_x$ of the belief function $\mathcal{H}_x$ is

$$m_x(F) = \sum \{P(\omega)/P(v_x(\Omega)) : \Gamma_x(\omega) = F\} \tag{7}$$

for all subsets $F \subseteq \Theta$. The logic behind the derivation of these equations is given in Monney [9].

From a GFM we can logically derive a unique distributional model as follows. For every $\theta \in \Theta$, let $P_\theta$ be the probability distribution on $X$ given by

$$P_\theta(x) = P(\{\omega \in \Omega : f(\theta, \omega) = x\}). \tag{8}$$

The plausibility function $Pl_x$ associated with the belief function $\mathcal{H}_x$ has a very interesting property. Indeed, if $l_x$ denotes the likelihood function of the distributional

model associated with the GFM when $x$ is observed, then it can easily be verified that

$$Pl_x(\theta) = c \cdot l_x(\theta) \tag{9}$$

for all $\theta \in \Theta$, where $c$ is a positive constant that does not depend on $\theta$. This means that the plausibility function restricted to simple hypotheses is proportional to the likelihood function of the associated distributional model. From this result, given an observation $x$, it is quite natural to define the weight of evidence of a general hypothesis $H \subseteq \Theta$ over a general hypothesis $H' \subseteq \Theta$ by

$$w_x(H, H') := Pl_x(H)/Pl_x(H'). \tag{10}$$

Then the weight of evidence of the simple hypothesis $H = \{\theta\}$ over the simple hypothesis $H' = \{\theta'\}$ is

$$
\begin{aligned}
Pl_x(\theta)/Pl_x(\theta') &= (c \cdot l_x(\theta))/(c \cdot l_x(\theta')) \\
&= l_x(\theta)/l_x(\theta'),
\end{aligned}
$$

which shows that the definition (10) extends to composite hypotheses the weight of evidence for simple hypotheses given in (3).

Of course, the concepts defined above for a single observation can easily be extended to the situation of several independent observations. First consider the case of distributional models. For all $\theta \in \Theta$, let $P_\theta$ be a distribution on $X$. In a total of $r$ independent observations of the variable $\xi$, suppose that the value $x$ is observed $m$ times and the value $x'$ is observed $n$ times $(m + n = r)$. These observations are considered as independent realisations of the variable $\xi$. Given these observations, the likelihood of the parameter value $\theta$ is

$$l_{m,n}(\theta) := \prod_{i=1}^{m} l_x(\theta) \cdot \prod_{i=1}^{n} l_{x'}(\theta). \tag{11}$$

Then the law of likelihood implies that the weight of evidence of $\theta$ over $\theta'$ is

$$
\begin{aligned}
w_{m,n}(\theta, \theta') &:= l_{m,n}(\theta)/l_{m,n}(\theta') \\
&= \prod_{i=1}^{m} w_x(\theta, \theta') \cdot \prod_{i=1}^{n} w_{x'}(\theta, \theta'). \tag{12}
\end{aligned}
$$

However, within the given distributional model, the weight of evidence for composite hypotheses is not defined.

Now consider the same observations from the point of view of a GFM having the given distributions $P_\theta$ as its associated distributional model. Since the observations are independent, it is reasonable to use Dempster's rule of combination to combine the belief functions coming from all the observations taken individually. The resulting belief function on $\Theta$, which is denoted by $\mathcal{H}_{m,n}$, represents the information generated on $\Theta$ based on all the observations made :

$$\mathcal{H}_{m,n} := (\oplus_{i=1}^m \mathcal{H}_x) \oplus (\oplus_{i=1}^n \mathcal{H}_{x'}). \qquad (13)$$

If $Pl_{m,n}$ denotes the plausibility function of $\mathcal{H}_{m,n}$, then it is natural to define the weight of evidence of a hypothesis $H \subseteq \Theta$ over a hypothesis $H' \subseteq \Theta$ by

$$w_{m,n}(H, H') := Pl_{m,n}(H)/Pl_{m,n}(H'). \qquad (14)$$

By theorem 4.9 of Kohlas & Monney [8], for all $\theta \in \Theta$ we have

$$Pl_{m,n}(\theta) = c \cdot (\prod_{i=1}^m l_x(\theta) \cdot \prod_{i=1}^n l_{x'}(\theta)) \qquad (15)$$

where $c$ is a positive constant which does not depend on $\theta$. This implies that the general definition of weight of evidence given in equation (14) generalizes to any type of hypothesis the weight of evidence for simple hypothesis given in equation (12) :

$$\frac{Pl_{m,n}(\theta)}{Pl_{m,n}(\theta')} = \prod_{i=1}^m w_x(\theta, \theta') \cdot \prod_{i=1}^n w_{x'}(\theta, \theta'). \qquad (16)$$

If the belief function $\mathcal{H}_{m,n}$ happens to be precise, i.e. all its focal sets are one-element subsets of $\Theta$, then it can be proved that

$$w_{m,n}(H, H') = \frac{\sum_{\theta \in H} l_{m,n}(\theta)}{\sum_{\theta \in H'} l_{m,n}(\theta)}. \qquad (17)$$

Another special case occurs when the belief function $\mathcal{H}_{m,n}$ is consonant, i.e. all its focal sets are nested. In this case we have

$$w_{m,n}(H, H') = \frac{\max \{l_{m,n}(\theta) : \theta \in H\}}{\max \{l_{m,n}(\theta) : \theta \in H'\}}. \qquad (18)$$

The last two equations are consequences of results on precise and consonant belief functions that can be found in Kohlas & Monney [8]. The situation where more than two different values $x$ and $x'$ are observed can be treated in a similar way.

# 3    Functional and Distributional Models

We have seen that we need a generalized functional model to compute the weight of evidence for composite hypotheses. The problem is that in general there exist several GFM having the same associated distributional model. This means that if the initial knowledge is given in the form of a distributional model, then the weight of evidence will depend on the GFM that is chosen to represent the given distributional model. To illustrate this fact, we are going to consider a particular distributional model and analyze two different GFM representing it. However, it should be mentioned that there is no reason why the initial knowledge should necessarily be given in the form of a distributional model. Any functional model that faithfully reflects the mechanism of the generation of the observed data can serve as an initial model.

Suppose that an urn contains four balls which are either black or white. We successively and randomly draw a ball from the urn and observe its color before it is placed back into the urn. Let $\xi$ denote the variable indicating the color of the ball drawn. Suppose that we draw a total of $r$ balls and that $m$ happen to be white whereas $n = r - m$ happen to be black. Based on these observations, what can be inferred on the correct value of the parameter variable $t$ indicating the number of white balls in the urn ? In particular, what is the weight of the evidence represented by the observations with respect to certain hypotheses ? Let $\Theta = \{0, 1, 2, 3, 4\}$ denote the set of possible values of the variable $t$ (the unknown number of white balls in the urn) and let $X = \{0, 1\}$ denote the set of possible values of the observable variable $\xi$ where 1 represents white and 0 represents black. The distributional model for this problem is, for all $\theta \in \Theta$,

$$P_\theta(x) = \begin{cases} \theta/4 & \text{if } x = 1 \\ 1 - \theta/4 & \text{if } x = 0. \end{cases} \qquad (19)$$

The classical binomial and Bayesian models (which requires an additional prior distribution on $\Theta$) can be used to make statistical inferences on $\Theta$. But if we want to compute weights of evidence for any kind of hypotheses we need functional models. In the following two subsections we are going to present two GFMs having $P_\theta$ as their associated distributional model.

## 3.1    A First GFM for the Urn Problem

The generalized functional model that will be presented in this subsection is inspired from the idea of conditional embedding proposed by Smets [14]. The method of conditional embedding is also described in Shafer [13]. Since we already have defined the variables $t$ and $\xi$, we still have to specify the variable $\epsilon$ and its distribution along with the function $f$ expressing what must necessarily be observed for each possible value of $t$ and $\epsilon$. We take the variable $\epsilon$ to be indicating a particular relation between $\Theta$ and $X$. More precisely, the set of possible values of $\epsilon$ is $\Omega = \{\varphi_1, \varphi_2, \ldots, \varphi_8\}$ where each $\varphi_i$ is the indicator function of a certain subset $S_i$ of $\Theta$. In other words, $\varphi_i(\theta) = 1$ if $\theta \in S_i$ and 0 otherwise. The eight sets $S_1, \ldots, S_8$ are

$$S_1 = \{4\}, \ S_2 = \{1,4\}, \ S_3 = \{2,4\}, \ S_4 = \{3,4\},$$
$$S_5 = \{1,2,4\}, \ S_6 = \{1,3,4\}, \ S_7 = \{2,3,4\},$$
$$S_8 = \{1,2,3,4\}.$$

The distribution of $\epsilon$ is defined to be

$$P(\varphi_1) = 3/32, \ P(\varphi_2) = 1/32, \ P(\varphi_3) = 3/32,$$
$$P(\varphi_4) = 9/32, \ P(\varphi_5) = 1/32, \ P(\varphi_6) = 3/32,$$
$$P(\varphi_7) = 9/32, \ P(\varphi_8) = 3/32.$$

The last piece of the GFM still missing is the function $f : \Theta \times \Omega \to X$. This function is defined by

$$f(\theta, \varphi_i) = \varphi_i(\theta). \tag{20}$$

It can easily be proved that the distributional model associated with this GFM is $P_\theta$ given in (19).

If we observe a white ball ($\xi = 1$) in one draw, then the corresponding belief function $\mathcal{H}_1$ on $\Theta$ has eight focal sets which are

$$\begin{aligned}
\Gamma_1(\varphi_i) &= \{\theta \in \Theta : f(\theta, \varphi_i) = 1\} \\
&= \{\theta \in \Theta : \varphi_i(\theta) = 1\} \\
&= S_i
\end{aligned} \tag{21}$$

for all $i = 1, \ldots, 8$. Of course, the $m$-value of $S_i$ is $P(\varphi_i)$. We can show that if we observe only white balls in the $r$ draws (i.e. $m = r$ and $n = 0$), then the focal sets of the resulting belief function $\mathcal{H}_{m,0}$ are again $S_1, \ldots, S_8$ and

$$\begin{aligned}
Pl_{m,0}(\{1,2\}) &= (1/4)^m + (1/2)^m - (1/8)^m \\
Pl_{m,0}(\{2,3\}) &= (3/4)^m + (1/2)^m - (3/8)^m \\
Pl_{m,0}(\{1,3\}) &= (3/4)^m + (1/4)^m - (3/16)^m.
\end{aligned}$$

This can be used to compute the weights of evidence.

The case where only black balls are observed can be treated in a similar way.

Now consider the situation where we draw $r \geq 2$ balls and $m \geq 1$ happen to be white and $n \geq 1$ happen to be black ($r = m + n$). If $\mathcal{H}_{0,n}$ denotes the belief function resulting from the observation of the $n$ black balls, then

$$\mathcal{H}_{m,n} = \mathcal{H}_{m,0} \oplus \mathcal{H}_{0,n} \tag{22}$$

is the belief function corresponding to the observation of the black and white balls. Its plausibility function satisfies

$$Pl_{m,n}(\{1,2\}) = \frac{(1/4)^m(3/4)^n - (1/8)^m(3/8)^n + (1/2)^r}{K}$$

$$Pl_{m,n}(\{2,3\}) = \frac{(1/4)^n(3/4)^m - (1/8)^n(3/8)^m + (1/2)^r}{K}$$

$$Pl_{m,n}(\{1,3\}) = \frac{(1/4)^n(3/4)^m + (1/4)^m(3/4)^n - (3/16)^r}{K}$$

for some positive constant $K$. Once again, this can be used to compute the weights of evidence as a function of $m$ and $n$.

## 3.2 A Second GFM for the Urn Problem

As in the previous model, let $t$ denote the parameter variable indicating the number of white balls in the urn and $\xi$ the observable variable indicating the color of the ball that is drawn (1 for white and 0 for black). In this model we assume that the four balls in the urn are numbered from 1 to 4. In addition, if there is $\theta$ white balls in the urn, we make the assumption that the white balls are numbered from 1 to $\theta$ and the black balls are numbered from $\theta + 1$ to 4. This is an important assumption that will permit us to easily specify a functional model. It is similar to the condition characterizing the so-called structures of the first kind introduced by Dempster [5, 6]. Since randomly drawing a ball in the urn is equivalent to randomly selecting the number of a ball, let $\epsilon$ denote the variable representing the selected number and let $\Omega = \{1, \ldots, 4\}$ denote the set of its possible values. Of course the distribution of $\epsilon$ is $P(\omega) = 1/4$ for all $\omega \in \Omega$. To complete the specification of the GFM we define the function $f : \Theta \times \Omega \to X$ by

$$f(\theta, \omega) = \begin{cases} 1 & \text{if } \omega \leq \theta \\ 0 & \text{otherwise} \end{cases} \tag{23}$$

because we necessarily observe a white ball if the randomly selected number $\omega$ is less than or equal to $\theta$.

Let's prove that the distributional model associated with this GFM is $P_\theta$ given in (19). If $x = 1$, then we have

$$P(\{\omega \in \Omega : f(\theta, \omega) = 1\}) = P(\{\omega : \omega \leq \theta\}) = \theta/4.$$

Similarly, if $x = 0$, then we have

$$P(\{\omega \in \Omega : f(\theta, \omega) = 0\}) = P(\{\omega : \omega > \theta\}) = 1 - \theta/4.$$

If we observe a white ball in one draw, then the corresponding belief function $\mathcal{H}_1$ on $\Theta$ has four focal sets which are

$$\begin{aligned}
\Gamma_1(\omega) &= \{\theta \in \Theta : f(\theta, \omega) = 1\} \\
&= \{\theta \in \Theta : \omega \leq \theta\} \\
&= \{\omega, \omega + 1, \ldots, 4\}
\end{aligned} \tag{24}$$

for all $\omega \in \Omega = \{1, 2, 3, 4\}$. The $m$-value of each of these focal sets is $1/4$. Note that $\mathcal{H}_1$ is a consonant belief function because its focal sets are nested.

We can show that if we observe only white balls in the $r$ draws, then the focal sets of the resulting belief function $\mathcal{H}_{m,0}$ are again $\Gamma_1(\omega)$ for all $\omega \in \Omega = \{1, 2, 3, 4\}$, which shows that $\mathcal{H}_{m,0}$ is also a consonant belief function. Moreover, the $m$-value of the focal sets of $\mathcal{H}_{m,0}$ is given by

$$
\begin{aligned}
m_{m,0}(\{1,2,3,4\}) &= (1/4)^m \\
m_{m,0}(\{2,3,4\}) &= (1/2)^m - (1/4)^m \\
m_{m,0}(\{3,4\}) &= (3/4)^m - (1/2)^m \\
m_{m,0}(\{4\}) &= 1 - (3/4)^m.
\end{aligned} \quad (25)
$$

Since $\mathcal{H}_{m,0}$ is consonant, then by theorem 3.7 of Kohlas & Monney [8] we have

$$
Pl_{m,0}(H) = \max \{Pl_{m,0}(\theta) : \theta \in H\} \quad (26)
$$

for all $H \subseteq \Theta$ and

$$
Pl_{m,0}(\theta) = l_1(\theta)^m = (\theta/4)^m \quad (27)
$$

for all $\theta \in \Theta$ because it can be shown that $c = 1$ in equation (15). In particular we obtain

$$
\begin{aligned}
Pl_{m,0}(\{1,2\}) &= (1/2)^m \\
Pl_{m,0}(\{2,3\}) &= (3/4)^m \\
Pl_{m,0}(\{1,3\}) &= (3/4)^m,
\end{aligned} \quad (28)
$$

which in turn can be used to compute the weights of evidence.

The case where only black balls are observed can be treated in a similar way.

Now consider the situation where we draw $r \geq 2$ balls and $m \geq 1$ happen to be white and $n \geq 1$ happen to be black $(r = m + n)$. If $\mathcal{H}_{0,n}$ denotes the belief function resulting from the observation of the $n$ black balls, then

$$
\mathcal{H}_{m,n} = \mathcal{H}_{m,0} \oplus \mathcal{H}_{0,n} \quad (29)
$$

is the belief function corresponding to the observation of the black and white balls. Its plausibility function satisfies

$$
\begin{aligned}
Pl_{m,n}(\{1,2\}) &= N_{12}/K \\
Pl_{m,n}(\{2,3\}) &= N_{23}/K \\
Pl_{m,n}(\{1,3\}) &= N_{13}/K,
\end{aligned}
$$

where

$$
\begin{aligned}
N_{12} &= (1/4)^m(3/4)^n + (1/2)^n((1/2)^m - (1/4)^m) \\
N_{23} &= (1/4)^n(3/4)^m + (1/2)^m((1/2)^n - (1/4)^n) \\
N_{13} &= (1/4)^m(3/4)^n + (1/4)^n((3/4)^m - (1/2)^m)
\end{aligned}
$$

and $K$ is a constant. This in turn can be used to compute the weights of evidence. Note that the plausibility degrees obtained in this model are different from those obtained in the model considered in the previous subsection, which implies that the weights of evidence are also different.

## 4    Evidence About a Survival Rate

In this section we are going to generalize the last model to the situation where there is an arbitrary number of balls in the urn. However, in order to illustrate the applicability of the theory developed in this paper, we analyze a real-world example that gives a concrete significance to this urn problem. It is based on a real clinical problem from the early 1980s. The problem was to generate information about the survival rate of newborn babies with a certain critical health problem using a new treatment called extracorporeal membrane oxygenation (Bartlett et al. [2], Ware [15], Begg [3], see also Royall [11]). It had been shown that the survival rate with the old treatment is about 20% and scientists where quite confident that the new treatment might have a survival rate of at least 80%. Based on the observation of the effect of the new treatment on a few babies, the problem is to gauge the impact of this evidence for a survival rate of more than 80% versus a survival rate of 20%.

This problem is modelled in the following way. Consider the population $S$ of all babies suffering from this specific health problem and let $L$ denote the subset consisting of those babies who would live with the new treatment. Assuming that there are $N$ babies in the entire population $S$ and $\theta$ babies in the subpopulation $L$, the survival rate is defined to be $\theta/N$. Of course the number of surviving babies $\theta$ is unknown. Let $t$ denote the variable indicating the number of babies who would survive with the new treatment. The set of possible values of $t$ is $\Theta = \{1, 2, \ldots, N\}$. Let $\xi$ denote the variable indicating the result of the new treatment on a baby in $S$. We define that $\xi$ is 1 when the baby lives and 0 otherwise, so that the set of possible values of $\xi$ is $X = \{0, 1\}$. Also, it is assumed that every baby in the population $S$ has the same chance of being observed.

Using this information, the following distributional model between $\Theta$ and $X$ can be defined : for all $\theta \in \Theta$,

$$
P_\theta(1) = \theta/N, \quad P_\theta(0) = 1 - (\theta/N). \quad (30)
$$

Given some observations, Royall [11] determines the weight of evidence for the hypothesis $H_2' = \{0.8 \cdot N\}$ over the hypothesis $H_1' = \{0.2 \cdot N\}$. Since both of these hypotheses are simple, the distributional model is sufficient to compute this value. However, here we want to determine the weight of evidence for the hypothesis $H_2 = \{\theta : \theta \geq 0.8 \cdot N\}$ over the hypothesis $H_1 = H_1'$. As we have seen, for this purpose we need a GFM having $P_\theta$ as its associated distributional model. First it is assumed that the babies in $S$ are numbered from 1 to $N$ in such a way that the babies in $L$ are numbered first. In other words, the babies in $L$ are numbered from 1 to $\theta$ and the babies in $S - L$ are numbered from $\theta + 1$ to $N$. Let $\epsilon$ denote the variable representing the number of a baby randomly selected from $S$ and let $\Omega = \{1, \ldots, N\}$ denote the set of its possible values. Of course $P(\omega) = 1/N$ for all $\omega \in \Omega$. Then we are in a position to define the function $f : \Theta \times \Omega \to X$ of the GFM by

$$f(\theta, \omega) = \begin{cases} 1 & \text{if } \omega \leq \theta \\ 0 & \text{otherwise.} \end{cases} \quad (31)$$

This functions gives the result of the new treatment on the baby number $\omega$ when there are $\theta$ babies in the population $L$. The distributional model associated with this GFM is clearly $P_\theta$ given in (30).

Suppose that we observe a baby who lives whith the new treatment. From this evidence, what is the belief function $\mathcal{H}_1$ on $\Theta$ that can be inferred ? It turns out that this is a consonant belief function whose focal sets are

$$\begin{aligned} \Gamma_1(\omega) &= \{\theta \in \Theta : f(\theta, \omega) = 1\} \\ &= \{\theta \in \Theta : \omega \leq \theta\} \\ &= \{\omega, \omega + 1, \ldots, N\} \end{aligned} \quad (32)$$

for all $\omega \in \Omega$. If $[r..s]$ represents the set of all integers between $r$ and $s$ (the limits are included), this shows that the focal sets of $\mathcal{H}_1$ are $F_i = [i..N]$ for all $i = 1, \ldots, N$ and the $m$-value of $F_i$ is $1/N$ for all $i = 1, \ldots, N$.

Now suppose that the new treatment is given to $m$ babies and that they all live. If $\mathcal{H}_{m,0}$ denotes the belief function on $\Theta$ induced by these observations, then it can be shown that its focal sets are again $F_i, i = 1, \ldots, N$, which shows that $\mathcal{H}_{m,0}$ is also a consonant belief function. But what is the $m$-function of this belief function ? For $i = 1, \ldots, N$, let $m_i^{(m)}$ denote the $m$-value of the focal set $F_i$ of $\mathcal{H}_{m,0}$. Then Dempster's rule of combination implies that

$$m_i^{(m)} = (1/N)(i \cdot m_i^{(m-1)} + \sum_{l=1}^{i-1} m_l^{(m-1)}) \quad (33)$$

for all $i = 1, \ldots, N$. Defining the $N \times N$ matrix

$$T = \begin{pmatrix} 1/N & 0 & \ldots & \ldots & \ldots & \ldots & 0 \\ 1/N & 2/N & 0 & \ldots & \ldots & \ldots & 0 \\ 1/N & 1/N & 3/N & 0 & \ldots & \ldots & \ldots \\ \ldots & \ldots & \ldots & \ldots & \ldots & \ldots & 0 \\ 1/N & \ldots & 1/N & 1/N & i/N & 0 & \ldots \\ \ldots & \ldots & \ldots & \ldots & \ldots & \ldots & 0 \\ 1/N & 1/N & 1/N & \ldots & \ldots & \ldots & N/N \end{pmatrix}$$

and the vector $m^{(m)} = (m_1^{(m)}, \ldots, m_N^{(m)})'$ (the prime means transpose), these equations can be written as

$$m^{(m)} = T \cdot m^{(m-1)}. \quad (34)$$

Note that the matrix $T$ has the remarkable property of being stochastic, i.e. each of its columns sums to 1 and all its entries are non-negative. Considering the focal sets $F_1, \ldots, F_N$ as the $N$ states of a Markov chain, this implies that $T$ is the transition matrix of a Markov chain $Z_t, t = 1, 2, \ldots$ with

$$P(Z_t = F_i) = m_i^{(t)}. \quad (35)$$

Of course, the initial distribution of the chain is

$$P(Z_1 = F_i) = m_i^{(1)} = 1/N \quad (36)$$

for all $i = 1, \ldots, N$. As a consequence of equation (34) we have

$$m^{(m)} = T^{m-1} \cdot m^{(1)}. \quad (37)$$

Therefore, to compute the $m$-function of $\mathcal{H}_{m,0}$ we have to compute the $(m-1)$-th power of the matrix $T$. Fortunately, the matrix $T$ has $N$ different eigenvalues that are all real and by the Jordan decomposition method we obtain

$$T = MLM^{-1} \quad (38)$$

where

$$M = \begin{pmatrix} -1 & 0 & \ldots & \ldots & \ldots & \ldots & 0 \\ 1 & -1 & 0 & \ldots & \ldots & \ldots & 0 \\ 0 & 1 & -1 & 0 & \ldots & \ldots & \ldots \\ \ldots & \ldots & \ldots & \ldots & \ldots & \ldots & \ldots \\ 0 & \ldots & 0 & 1 & -1 & 0 & \ldots \\ \ldots & \ldots & \ldots & \ldots & \ldots & \ldots & \ldots \\ 0 & \ldots & \ldots & \ldots & 0 & 1 & 1 \end{pmatrix}$$

and

$$L = \begin{pmatrix} 1/N & 0 & \ldots & \ldots & \ldots & \ldots & 0 \\ 0 & 2/N & 0 & \ldots & \ldots & \ldots & 0 \\ 0 & 0 & 3/N & 0 & \ldots & \ldots & \ldots \\ \ldots & \ldots & \ldots & \ldots & \ldots & \ldots & \ldots \\ 0 & \ldots & \ldots & 0 & i/N & 0 & \ldots \\ \ldots & \ldots & \ldots & \ldots & \ldots & \ldots & \ldots \\ 0 & \ldots & \ldots & \ldots & \ldots & 0 & 1 \end{pmatrix}.$$

This implies that

$$T^{m-1} = M(L^{m-1})M^{-1} \quad (39)$$

and since $L^{m-1}$ is a diagonal matrix whose $(i,i)$-th element is $(i/N)^{m-1}$, a little algebra shows that

$$m_i^{(m)} = \frac{i^m - (i-1)^m}{N^m} \quad (40)$$

for all $i = 1, \ldots, N$.

When we observe that more and more babies survive with the new treatment and no one dies, i.e. when $m$ tends to infinity, the $m$-function of $\mathcal{H}_{m,0}$ tends to the $m$-function $m^*$ given by $m^*(\{N\}) = 1$ and $m^*(F_i) = 0$ for the other focal sets $F_i$. But $m^*$ represents the belief function asserting that $N$ is surely the exact value of the parameter, which means that every baby will survive with the new treatment. Of course, this result is compatible with our intuition because if we observe more and more babies who survive with new treatment (and never observe a baby who dies), then we are more and more inclined to believe that the new treatment will cure all babies. This asymptotic behavior of the belief function $\mathcal{H}_{m,0}$ is also a consequence of the fact that the state $\{N\}$ is the only closed class of the Markov chain, i.e. it is an absorbing state of the chain.

Since $\mathcal{H}_{m,0}$ is a consonant belief function and

$$Pl_{m,0}(\theta) = (\theta/N)^m \qquad (41)$$

for all $\theta \in \Theta$, it is easy to compute the degree of plausibility of any hypothesis $H \subseteq \Theta$ because

$$Pl_{m,0}(H) = \max \{Pl_{m,0}(\theta) : \theta \in H\}. \qquad (42)$$

This in turn allows us to compute the weight of evidence for any pair of hypotheses. For example, for the specific hypotheses $H_1$ and $H_2$ defined above we obtain

$$w_{m,0}(H_2, H_1) = \frac{1}{0.2^m} = 5^m. \qquad (43)$$

Now consider the situation where we observe a baby who does not survive with the new treatment. From this observation, what is the belief function $\mathcal{H}_0$ on $\Theta$ that can be inferred ? It turns out that this is a consonant belief function whose focal sets are $G_i = [0..(N-i)]$ for all $i = 1, \ldots, N$ and the $m$-value of $G_i$ is $1/N$ for all $i = 1, \ldots, N$.

Now suppose that the new treatment is given to $n$ babies and that none of them survive. If $\mathcal{H}_{0,n}$ denote the belief function on $\Theta$ induced by these observations, then the focal sets of $\mathcal{H}_{0,n}$ are again $G_i, i = 1, \ldots, N$, which shows that $\mathcal{H}_{0,n}$ is again a consonant belief function. But what is the $m$-function of this belief function ? If $m_i^{(n)}$ denotes the $m$-value of the focal set $G_i$ of $\mathcal{H}_{0,n}$, the Jordan decomposition method can be used to prove that

$$m_i^{(n)} = \frac{i^n - (i-1)^n}{N^n} \qquad (44)$$

for all $i = 1, \ldots, N$ because it is again possible to identify an underlying Markov chain. As above, this result can be used to compute $\mathcal{H}_{0,n}$ when $n$ tends to infinity.

Since $\mathcal{H}_{0,n}$ is a consonant belief function and

$$Pl_{0,n}(\theta) = (1 - (\theta/N))^n \qquad (45)$$

for all $\theta \in \Theta$, it is easy to compute the degree of plausibility of any hypothesis $H \subseteq \Theta$ because

$$Pl_{0,n}(H) = \max \{Pl_{0,n}(\theta) : \theta \in H\}. \qquad (46)$$

This in turn can be used to compute the weight of evidence for any pair of hypotheses. For example, we obtain

$$w_{0,n}(H_2, H_1) = \frac{0.2^n}{0.8} = 0.25 \cdot 0.2^{n-1}. \qquad (47)$$

Finally, let's suppose that among $r \geq 2$ babies to which the new treatment is administered, a total of $m \geq 1$ live and a total of $n \geq 1$ die $(m + n = r)$. Let

$$\mathcal{H}_{m,n} = \mathcal{H}_{m,0} \oplus \mathcal{H}_{0,n} \qquad (48)$$

denote the belief function on $\Theta$ induced by these observations. In this case, it turns out that for all $(i, j)$ in

$$\Delta = \{(i, j) \in \Omega \times \Omega : i + j \leq N\} \qquad (49)$$

the set $E_{ij} = [i..(N-j)]$ is a focal set of $\mathcal{H}_{m,n}$ and its $m$-value is

$$m_{m,n}(E_{ij}) = \frac{(i^m - (i-1)^m)(j^n - (j-1)^n)}{K} \qquad (50)$$

where

$$
\begin{aligned}
K &= \sum_{l=1}^{N} (l^n - (l-1)^n)(N-l)^m \\
&= \sum_{l=1}^{N} (l^m - (l-1)^m)(N-l)^n.
\end{aligned}
$$

Note that the belief function $\mathcal{H}_{m,n}$ is no longer consonant. If $Pl_{m,n}$ denotes its plausibility function, then we can easily find that

$$Pl_{m,n}(\theta) = \frac{\theta^m (N-\theta)^n}{K} \qquad (51)$$

for all $\theta \in \Theta$. In addition, the degree of plausibility of $[0..r]$ and $[1..r]$ are the same, and the degree of plausibility of $[r..N]$ and $[r..(N-1)]$ are the same. Finally, it can also be verified that

$$
\begin{aligned}
Pl_{m,n}([1..r]) &= \frac{\sum_{l=1}^{r}(l^m - (l-1)^m)(N-l)^n}{K} \\
Pl_{m,n}([r..N]) &= \frac{\sum_{l=1}^{N-r}(l^n - (l-1)^n)(N-l)^m}{K}
\end{aligned}
$$

for all $r$ in $\{0, \ldots, N\}$ and

$$Pl_{m,n}([r..s]) = Pl_{m,n}([1..s]) + Pl_{m,n}([r..N]) - 1 \qquad (52)$$

for all $r$ and $s$ such that $0 \leq r \leq s \leq N$. These results can be used to compute the weight of evidence for any pair of hypotheses. For example, we obtain

$$w_{m,n}(H_2, H_1) = \frac{\sum_{l=1}^{0.2 \cdot N}(l^n - (l-1)^n)(N-l)^m}{0.2^m \cdot 0.8^n \cdot N^r}.$$

If we collect the actual observations resulting from two studies about the effectiveness of the new treatment, we find that among $r = 40$ babies who received the new treatment a total of $m = 39$ babies survived and only $n = 1$ baby died. In this case we get

$$w_{39,1}(H_2, H_1) = \frac{\sum_{l=1}^{0.2 \cdot N} (N - l)^{39}}{0.2^{39} \cdot 0.8 \cdot N^{40}}. \quad (53)$$

A finer analysis of this function of $N$ reveals that it stays approximately constant for $N \geq 250$ :

$$w_{39,1}(H_2, H_1) \approx 5 \cdot 10^{25} \quad (54)$$

for $N \geq 250$. This means that the weight of the available evidence for $H_2$ versus $H_1$ is approximately $w = 5 \cdot 10^{25}$. In order to give an intuitive interpretation to this value, consider an urn containing two balls which can be either black or white. If $WW$ denotes the hypothesis that they are both white and $BW$ denotes the hypothesis that there is one black and one white, then the weight of evidence $w = 5 \cdot 10^{25}$ is approximately the same as the weight of evidence for $WW$ versus $BW$ when we observe 85 white balls in 85 random draws of a ball from the urn. This is pretty strong evidence for the hypothesis that the new treatment's survival rate is at least 80% versus the hypothesis that it is 20% (the old treatment's survival rate).

## 5    Conclusion

We have defined a general concept of weight of evidence that is applicable to any kind of parametric hypothesis. Classical distributional models do not convey enough information to generate a natural and general concept of weight of evidence. In other words, they are too coarse a representation of the mechanical process underlying the generation of the data observed. Generalized functional models and the theory of belief functions are the appropriate tools for defining such a general concept. Finally, a weight of evidence can be given a concrete significance by finding a well understood and simple situation leading to the same weight of evidence.